# Similar Image Search for Histopathology: SMILY

Narayan Hegde[1*], Jason D. Hipp[1*], Yun Liu[1], Michael Emmert-Buck[2] ,Emily Reif[1], Daniel Smilkov[1], Michael Terry[1], Carrie J. Cai[1], Mahul B. Amin[3], Craig H. Mermel[1], Phil Q. Nelson[1], Lily H. Peng[1], Greg S. Corrado[1], Martin C. Stumpe[1]

## Abstract

The increasing availability of large institutional and public histopathology image datasets is enabling the searching of these datasets for diagnosis, research, and education. Though these datasets typically have associated metadata such as diagnosis or clinical notes, even carefully curated datasets rarely contain annotations of the location of regions of interest on each image. Because pathology images are extremely large (up to 100,000 pixels in each dimension), further laborious visual search of each image may be needed to find the feature of interest. In this paper, we introduce a deep learning based reverse image search tool for histopathology images: Similar Medical Images Like Yours (SMILY). We assessed SMILY's ability to retrieve search results in two ways: using pathologist-provided annotations, and via prospective studies where pathologists evaluated the quality of SMILY search results. As a negative control in the second evaluation, pathologists were blinded to whether search results were retrieved by SMILY or randomly. In both types of assessments, SMILY was able to retrieve search results with similar histologic features, organ site, and prostate cancer Gleason grade compared with the original query. SMILY may be a useful general-purpose tool in the pathologist's arsenal, to improve the efficiency of searching large archives of histopathology images, without the need to develop and implement specific tools for each application.

[1] Google AI Healthcare, Mountain View, CA 94043, USA
[2] Avoneaux Medical Institute, Baltimore, MD 21215, USA
[3] Department of Pathology and Laboratory Medicine, University of Tennessee Health Science Center, Memphis, TN 38163, USA
[*]Equal Contributions

# Introduction

The growing adoption of digital pathology[1] provides opportunities to archive and search large databases of pathology images for diagnosis, research, and education. Histopathology is the examination of biological tissue specimens for diagnostic purposes and is traditionally performed using microscopes. After digitization, images "tagged" (annotated) with clinical data such as diagnoses and patient demographics can be searched based on the text-based tags. For example, searching for "breast" and "carcinoma" in the clinical notes could yield a list of images that were diagnosed or suspected to contain breast cancer.

A relatively unique aspect of histopathology images is that they are typically much larger than those found in other imaging specialties: a typical pathology slide might be 100,000×100,000 pixels when digitized at high magnification. Since clinical annotations such as text reports apply to the entire image or sets of images rather than specific locations within the image, matching a search "query" with the location in the image that the search is relevant to can be challenging. For instance, a tumor in a pathology image may be only 100 pixels across, comprising *one-millionth* of the image area. A clinician, researcher, or trainee who has found this image or set of images via searching based on text would still need to visually search the image to locate the lesion before any subsequent analysis. This problem is further compounded because like many disciplines, real-world pathology cases contain multiple (e.g. 5-100) images, and the available text labels might not be specific enough in terms of a particular disease subtype of interest.

In non-medical domains, a potential solution is *reverse image search*, also termed content-based image retrieval (CBIR)[2], to find visually "similar" images. In the diagnostic workflow for example, a clinician may want to search a database for similar lesions to determine if a feature of interest is malignant or a benign histologic mimic, for example in basal cell carcinoma[3]. Relevant tools in non-medical domains include "search by image" for general images[4], visual search[5] for retail products, and other tools for faces[6] and art[7]. In medical imaging, related works include CBIR for radiology[8–10] and pathology[11–18]. Prior machine learning based CBIR systems have employed application-specific models, which require collecting labeled data for each application, creating a significant burden to their implementation. Furthermore, "similarity" in these works were defined along specific axes, whereas the intended meaning could vary based on the use case. For example, two images could be similar in that they originate from the same organ, same cancer, similar staining, or similar histologic features.

In this paper, we developed a histopathology similar image search tool (Similar Medical Images Like Yours, SMILY) without using labeled histopathology images. We then evaluated histopathology image search quality in several organs: breast, prostate, and colon, representing three of the four most common non-cutaneous cancer sites. Our evaluation had two components. First, we quantitatively evaluated how often a query image would be matched to an appropriate result from a dataset that was pre-annotated by pathologists. Second, in a blinded prospective study, we had pathologists evaluate how a query image compared to search results from a dataset that were selected either using SMILY or a random image patch. In both evaluations, we assessed SMILY's ability to retrieve similar tissue types, histologic features and even disease states such as prostate cancer Gleason grading.

# Results

Fig. 1 provides an overview of the development and usage of our proposed tool, SMILY. The first step is to create the SMILY database for searching. The core algorithm of this step is a convolutional neural network that condenses image information into a numerical feature vector, termed an embedding. When computed across image patches cropped from slides, this created a database of patches and a numerical summary of each patch's information content. When a region of interest is selected for searching (termed the query image), the embedding for the query image is computed and compared with those in the database to retrieve the most similar patches (Methods). An example of the user interface for SMILY is presented in Fig. 2. In the following evaluations, the database was constructed using images at medium (10X) magnification from the publicly available TCGA (The Cancer Genome Atlas)[19]. In total, the evaluations used 127,000 image patches from 45 slides and the query set consisted of 22,500 patches from another 15 slides.

Labels for the first type of evaluation were prepared by pathologists annotating various histologic features in the images, such as arteries, nerves, smooth muscle, and fat. Despite non-exhaustive annotations, this process produced a total annotated area exceeding 150,000 8×8 μm regions (each roughly equivalent to a lymphocyte). After sampling to ensure a balanced dataset with respect to classes of interest in each analysis, this produced thousands of image patches per class (Methods, Table 1). Next, for each query patch, we evaluated the performance of SMILY in retrieving patches of the same histologic feature in the database. We used the top-5 score, which evaluates the ability of SMILY to correctly present at least one correct result in the top 5 search results. This metric was chosen to mimic the standard search process, where a user evaluates a small number of search results to find matches of interest. The subsequent evaluations used image patches extracted at 10X "medium power" magnification, which is commonly used for reviewing images. The results on other magnifications and using other performance metrics are presented in Supp Figs. 1 and 2, respectively.

Fig. 3A illustrates the results of this large-scale quantitative analysis. When we used query images from prostate specimens, SMILY had a 63.9% top-5 score at retrieving images of the same histologic feature. This was significantly higher than a traditional image feature extractor (scale-invariant feature transform, SIFT) used in related work[17] (45.2%, $p<0.001$) and random (28.3%). When SMILY retrieved results that did not exactly match the histologic feature, it commonly returned a similar feature, such as another fluid-transporting vessel: capillaries, arteries, veins, and lymphatics (Fig. 4A). Next, we expanded to queries from multiple organs: breast, colon, and prostate. For most histologic features, errors tended to occur between the same histologic features, but across organs (Fig. 4B). For example, the histologic feature match score was at 65.3%, but the combined histologic feature and organ match was lower at 40.0%. Finally, we evaluated the ability of SMILY to retrieve images of the same prostate cancer Gleason pattern (Fig. 3B). SMILY was significantly more accurate than the SIFT baselines at retrieving images with the correct Gleason patterns (76.0% vs. 65.2%, $p<0.001$), and frequently with both the same Gleason pattern and the same histologic feature (25.3% vs. 17.8%, $p<0.001$ for comparison with SMILY).

The previous analyses used a large number of patches for evaluations: more than 20,000 patches in the query set and 5 times that number in the database. However, one limitation was that they were based on non-exhaustive annotations of histologic features and Gleason patterns. For example, if a query image contains fat only, a retrieved image search result that contains both fat and an artery (but is only annotated as “artery”) will be considered an error during the prior evaluation. Thus, our second set of evaluations involved a study with pathologists to assess if at least one histologic feature or cancer grade present in the query image exists in the SMILY search results. As a control to ensure that graders were not artificially scoring SMILY results highly, some search results were from a random search instead of SMILY. The graders were blinded to the source of the search results: SMILY or random. Analogous to the large-scale quantitative stores above, we then assessed search results along multiple axes: histologic feature, organ site, and Gleason grading (Table 2).

Using queries from prostate specimens, SMILY had an average score (Methods) of 62.1% for finding similar histologic features, significantly higher than the random search results (26.8%, $p<0.001$) (Fig. 5A). “Random” performance exceeds the inverse of the number of categories (1/9=11%) because each search result can contain multiple histologic features. When we queried from multiple organs, SMILY’s score for histologic feature match was similarly significantly higher than random (57.8% vs 18.3%, $p<0.001$, Fig. 5B). In this study, pathologists reported the organ site as unambiguous for only 32.0% of the individual search results. Among these search results, 68.3% were from the same organ site as the query image (Fig. 5C). Finally, graders provided a 0-100 “match quality score” for prostate cancer patches, based on tumor presence, Gleason pattern and histologic features (Methods and Table 3). In this analysis, SMILY scored an average of 61.0%, compared with 30.0% for random results (Fig. 5D, $p<0.001$).

To better understand SMILY, we visualized the embeddings using t-SNE, a common tool for understanding where data lie in a high-dimensional embedding space[20]. Fig. 6A shows that image patches from the same organ site can lie in very different areas in the embedding space. When colored based on histologic features, the clusters tend to have more distinct colors (Fig.

6B). For example, the bottom left prostate cluster in Fig 6A is composed of a mixture of histologic features such as arteries, lymphatic vessels and capillaries in Fig 6B.

Finally, to investigate the computational efficiency of SMILY for datasets of realistic sizes, we created a database for all prostate, breast, lung, and colon specimens from TCGA, at 4 magnifications: 40X, 20X, 10X, and 5X. This generated about $10^9$ image patches. Using 400 computers with 10 compute threads each, and some optimizations to use a hash table instead of the kd-tree depending on the local embedding density[21], queries had a median query time of 1.3 seconds. This can in principle be further accelerated using text-based search to filter images, and real-time updating of search results to present preliminary results before the search completes. By contrast, a naive implementation on a single machine with $10^7$ image patches (100 times fewer than above) required a significantly slower 25 seconds per query.

# Discussion

This study presents SMILY, a tool to search for similar histopathology images using an image as the query. To our knowledge, we have performed the most comprehensive evaluation of a reverse image search tool for histopathology. SMILY retrieves image search results with similar histologic features, organ site, and cancer grades, based on both large-scale quantitative analysis using annotated tissue regions and prospective studies with pathologists blinded to the source of the search results. In the rest of this discussion, we will discuss some nuanced issues regarding similar image search: what 'similarity' means; what a tool like SMILY can be used for; comparison with "traditional" application-specific approaches; how SMILY was developed and what that means for future applications not covered in our evaluations; comparison with prior work; and finally technical implementation considerations.

First, the meaning of "accuracy" in the setting of a similar image search tool deserves some thought. From first principles, the ideal search tool displays what you are searching for. However, this goal is ambiguous because the intent of the search depends on the use case: searching for other images with the same stain, similar stain intensity, same histologic feature, or similar lesion in the most general sense. As such, in the absence of information about search intent, the ideal tool should surface a breadth of search results instead of focusing on any single axis of similarity. To address the lack of algorithmic awareness for the search intent, advances in human-computer interaction may enable interactive refinement of search results based on certain desired axes of similarity[22].

The potential use cases for a tool like SMILY can be categorized into diagnosis, research, and education. In diagnosis, SMILY could be a helpful tool to search for similar lesions within the same slide or in other patients. For example, when a lesion or histologic feature (e.g. mitotic figure, prominent nucleoli, necrosis, etc) is found, the pathologist might want to know the frequency of that lesion's or histologic feature's occurrence in the specimen. For example, counting the frequency of mitosis is clinically relevant in breast cancer[23]. Searching for rare features in other slides may be helpful in rare diagnoses, to better understand the prognosis of other patients with similar features, including potentially rare pathologies from historic cases with known intervention and treatment response. For research, a clinician might have a hypothesis: occurrence of a certain histopathological feature in the slide is correlated with clinical endpoints. However, an adequately powered study may require a large number of patients, rendering the manual search for these features highly labor intensive. SMILY could enable significant speedups in this search via computer-assisted search. Finally, trainees are frequently confronted with unknown lesions. Manual searching of pathology textbooks, atlases, and other resources for similar lesions can be time-consuming; SMILY could reduce this process to an image-based query and manual assessment for the most relevant result. Importantly, these searches could also leverage large publicly available databases such as the TCGA, as we have done here.

Indeed, with respect to specific applications such as mitotic counting[24], approaches that have been developed specifically for that application may result in higher accuracy for that purpose. However, developing and implementing specific but separate approaches for every possible task of interest is impractical. Some challenges are: expensive data collection and labeling, difficulty of workflow integration and potential legal or commercial issues, and lack of machine learning, software or hardware expertise for development or implementation. As such, the availability of a general-purpose tool like SMILY that can be used in multiple applications, can be helpful despite having lower accuracy than an application-specific tool.

An interesting aspect of SMILY is that the core neural network algorithm was not trained using histopathology images. Instead, the network was trained using a dataset of images including people, animals, and man-made and natural objects (see Methods). Thus our approach does not require the use of large, pixel-annotated datasets such as those used for breast cancer mitotic figure detection[25], breast cancer metastasis[26], or “image-level” labels such as those extracted from pathology reports[27]. Because the network is not optimized for the image characteristics of any given laboratory or slide scanner, the network is likewise not overfit to these aspects, and can be expected to generalize to other datasets. In principle, the development of a similar histopathology “similarity” dataset could further improve the embeddings learned by the model, and is the subject of future work. The results of using several other “pre-trained” neural networks are presented in Supp Fig. 4.

CBIR has been studied extensively in medical imaging[9,10,28], and in histopathology for both slides[14,18] and image patches[11–13,15,29]. However, the models underlying these CBIR systems require pathologist-annotated labels for development, which is both costly and non-scalable. These annotations also in turn restricts the concept of similarity to be along a few predetermined axes, such as cancer grade and histologic features. In prior work, lookup accuracies ranged from 60-80% for breast, prostate, necrotic and leukocyte organ sites and histologic features[13,15]. By contrast, SMILY achieved comparable performance without the use of application-specific labeled data for development (though we collected annotations for evaluation purposes), and thus can be applied to applications without labeled training data. Alternative histopathology CBIR approaches in the absence of labeled training data include SIFT, kernel and Fourier functions[17,30]. We have performed a quantitative comparison with SIFT to evaluate the added value of using a neural-network-based system like SMILY for automatic feature extraction, showing a significant improvement in lookup accuracy across multiple image retrieval experiments.

The large size of each histopathology image and the scale of typical histopathology databases ($10^3$-$10^6$ images) raise important technical considerations for real world use. First, the embedding of each patch needs to be calculated as a one-time computation cost. This incurs a delay to compute the embeddings for the $10^3$-$10^6$ patches in each newly digitized slide before the slide can be searched across. Second, these embeddings need to be stored to avoid repeated embedding recomputation. Though this overhead was only 0.4% per gigabyte-sized image in our studies, this storage requirement increases with the number of magnifications of interest, and density of sampling each slide. Finally, the search phase requires comparing the query image embedding with millions or billions of other embeddings. For example, a naive implementation of this process on the entirety of the publicly-available The Cancer Genome Atlas (TCGA, contains over 33,000 slides) dataset[31] will incur an impractical, half-minute latency on a modern desktop computer. To support real-world usage, we have optimized this process to require only seconds on a web interface (Methods).

This study contains limitations, such as those discussed in-depth above regarding accuracy of a similar image search tool and limitation of a general-purpose search tool versus application specific tools. In addition, the number of slides could be increased to better capture the breadth of tissue processing conditions and resulting images. We have evaluated ‘similarity’ in terms of histologic features, organ site, and prostate cancer grade, but there are other axes of similarities that SMILY will need to be further validated on. Lastly, future work will also need to tackle the ‘refinement’ of search results along specific similarity axes of interest, to enable more targeted image-based search for histopathology.

# Methods

## The SMILY tool

### Neural network architecture

SMILY is based on a convolutional neural network architecture called a deep ranking network[32]. Briefly, the deep ranking network is based on an embedding-computing module that compresses input image patches (of dimensions width x height x channels) into a fixed-length vector. This module contains layers of convolutional, pooling, and concatenation operations. During training, the network was fed labeled sets of 3 images: a reference image $I$ of a certain class, a second image $I_+$ of the same class, and a third image $I_-$ of a different class. The network then uses the modules to compute the embeddings of each of the 3 images. The network is then trained to assign a lower distance between the embeddings of ($I$, $I_+$) than the embedding distances of ($I$, $I_-$). Our network was trained on about 500,000,000 "natural images" (e.g. dogs, cats, trees, man-made objects etc) from 18,000 distinct classes. In this way, the network learned to distinguish similar images from dissimilar ones by computing and comparing the embeddings of input images.[32,33]

## Building the SMILY embedding database

For the experiments described in this paper, we used slides from The Cancer Genome Atlas (TCGA)[19]. TCGA was used because it is publicly available and widely used for histopathology studies. TCGA tissue samples were collected with approval of local Institutional Review Boards (IRBs), with the informed consent of patients. Ethics review and IRB exemption for the use of de-identified images in this study was obtained from Quorum Review IRB (Seattle, WA).

Additional details about each experiment's dataset are provided in the respective study sections. SMILY uses the embedding-computing module from the deep ranking network (Fig. 1) to compress input image patches (300×300 pixels 3-channel RGB (red-green-blue) images) into embeddings vectors of size 128. Because histopathology images are orientation-independent, we additionally generate the four 90-degree rotations of each input image, and the mirrored and rotated versions for a total of 8 orientations, and correspondingly eight 128-sized embeddings per image patch, a 260-fold dimensionality reduction. Even in the absence of any additional compression, storing these embeddings required a reasonable additional 0.4% storage overhead compared to storing the original images alone.

To create image patches at a given image magnification (e.g. high magnifications like 40X and 20X, or medium magnification like 10X), we extracted thousands of non-overlapping patches per category (Table 1). For a real use case, overlaps can be used to ensure each histologic feature is contained entirely in a patch of the appropriate magnification, instead of being potentially bisected into two patches.

## Querying the database

To retrieve matches from the database, SMILY first computes the embedding for a selected query image patch, and then compares that embedding with the embeddings stored in the database. For this work, our comparison function was the $L_2$ distance across pairs of 128-sized embedding vectors. To handle the 8 orientations (see above), we filtered the search results such that the most similar orientation was returned, and only one orientation for each distinct image patch was presented in each set of search results. In addition, to enhance diversity of the search results, we filtered the results to ensure that no results were within 1,000 pixels of each other.

Our experiments (described below) required large numbers of embedding comparisons, ranging from 40,000 to 90,000. To enable efficient lookups, we used k-dimensional (k-d) trees[34] with a leaf size of 40 and depth of 6; this is customizable to fit computation resources and speed requirements. To further optimize lookup speed, we parallelized the comparisons across multiple machines (Results). These steps provided a lookup time sublinear in the number of comparisons.

### SMILY's user interface

SMILY was implemented as a web-based whole-slide viewer (Fig. 2). To conduct a search, a user selects a rectangular image patch between 200 and 400 pixels in height and width. For a query patch that is not 300×300 pixels, SMILY resizes to 224×224 pixels using bilinear interpolation before computing the embedding. The embedding is then used to search the database based on the current magnification in the selected region, and the results are displayed as a customizable number of image patches. Optionally, any existing pixel-level annotations or slide-level metadata such as the original diagnosis can be displayed as well.

## Evaluations

To evaluate the utility of SMILY, we conducted several experiments by building a SMILY database using the TCGA dataset, and then conducting studies to examine the quality of SMILY image search results.

### Large-scale quantitative studies

Our large-scale quantitative experiments were based on regions annotated by pathologists with various labels (Table 1). In each case, pathologists annotated slides with various histologic features (up to 11 categories) or Gleason patterns (4 categories: non-tumor and 3 Gleason patterns). Annotations were performed by three pathologists using the Hamamatsu NDPview2 whole-slide image viewer [35], using the free-hand outlining and labeling tool. These annotations were used only for evaluating SMILY, and not for developing the SMILY embedding neural network. Because of the large size of each slide and the complexity in the appearances of each feature of interest, annotations were not required to be exhaustive.

Patches of size 300×300 pixels for each histologic feature or Gleason pattern category were then extracted based on the annotations and stored in the SMILY database along with their embeddings computed at the appropriate magnification. To avoid class imbalance for these experiments, we subsampled hundreds to thousands of patches without replacement for each category (Table 1).

### Prospective studies with pathologists

In addition to the large-scale studies, we conducted similar studies by asking pathologists to rate the quality of search results. Because our annotations (for the large-scale quantitative studies) were non-exhaustive, these studies with pathologists allowed regions containing multiple labels (but annotated only as one label) to be assessed correctly. The same database from the large-scale quantitative studies were used, but the query data were subsampled to hundreds instead of thousands, to retain a tractable number of search results for manual evaluation by pathologists. To mimic the use case of a user assessing multiple search results for a single query, each query generated 4 search results, and the pathologist rated each of the 4 results (the scoring system for each experiment is described below). The final average score for each study is the average score across all search results for all queries.

Similarity along histologic feature, organ, and Gleason pattern were assessed analogously to the large-scale studies using binary scores (100 for “match” and 0 for “not-match”). For an overall “match quality score” combining multiple axes, we devised a 100-point score in 25-point increments (Table 3). A few samples of the user interface for the histologic feature and organ similarity assessment is presented in Supp Fig. 3.

In total, 3 anatomic pathologists from diverse backgrounds participated in this study: 1 U.S. board-certified, 1 non-U.S. board-certified, and 1 U.S. residency-trained. As a negative control to ensure that our pathologists were not artificially rating SMILY search results highly, 25% of the queries returned search results from random selection (i.e. all 4 images were from SMILY or all 4 images were randomly selected). The pathologists were blinded to the source of the search results: SMILY versus random.

### Statistical analysis

To assess the statistical significance of our results, we used a two-tailed Chi-squared test. Because of the large size of each study, most differences were statistically significant.

## Acknowledgements

For technical advice and discussion, we thank the following, who are all employees of Alphabet Inc: Alvin Rajkomar, MD, Daniel Tse, MD. For software infrastructure, logistical support, and slide digitization services, we thank members of the Google AI Healthcare Pathology team. Lastly, we are deeply grateful to the pathologists Isabelle Flament, Trissia Brown, and Stanislaw Krajewski who provided annotations for this study, or participated in the “prospective study with pathologists”.

## Author Contributions

L.H.P. and M.C.S. conceived the idea; N.H., J.D.H., Y.L., and M.T. designed the experiments; N.H., E.R., D.S., and C.J.C. wrote code for various parts of the work; L.H.P., M.C.S., P.Q.N., and G.S.C. acquired the tissue samples for use in the study and provided strategic support; M.B.A., M.E.B., and J.D.H. created the guidelines for the studies with pathologists and designed the evaluation scoring systems; C.H.M. and M.C.S. supervised the project; N.H., Y.L., C.H.M., and J.D.H. wrote the manuscript with the assistance and feedback of all other co-authors.

## Competing Interests

N.H., J.D.H., Y.L., E.R., D.S., M.T., C.J.C., C.H.M., P.Q.N., L.H.P., G.S.C., and M.C.S. are employees of Google LLC and own Alphabet stock. M.E.B. and M.B.A. were compensated for their expertise and time as pathologists. The authors declare no competing non-financial interests.

## Data Availability

Our study used images accessible from the Genomic Data Commons portal[36], which is based upon data generated by the TCGA Research Network[31].

# Figures

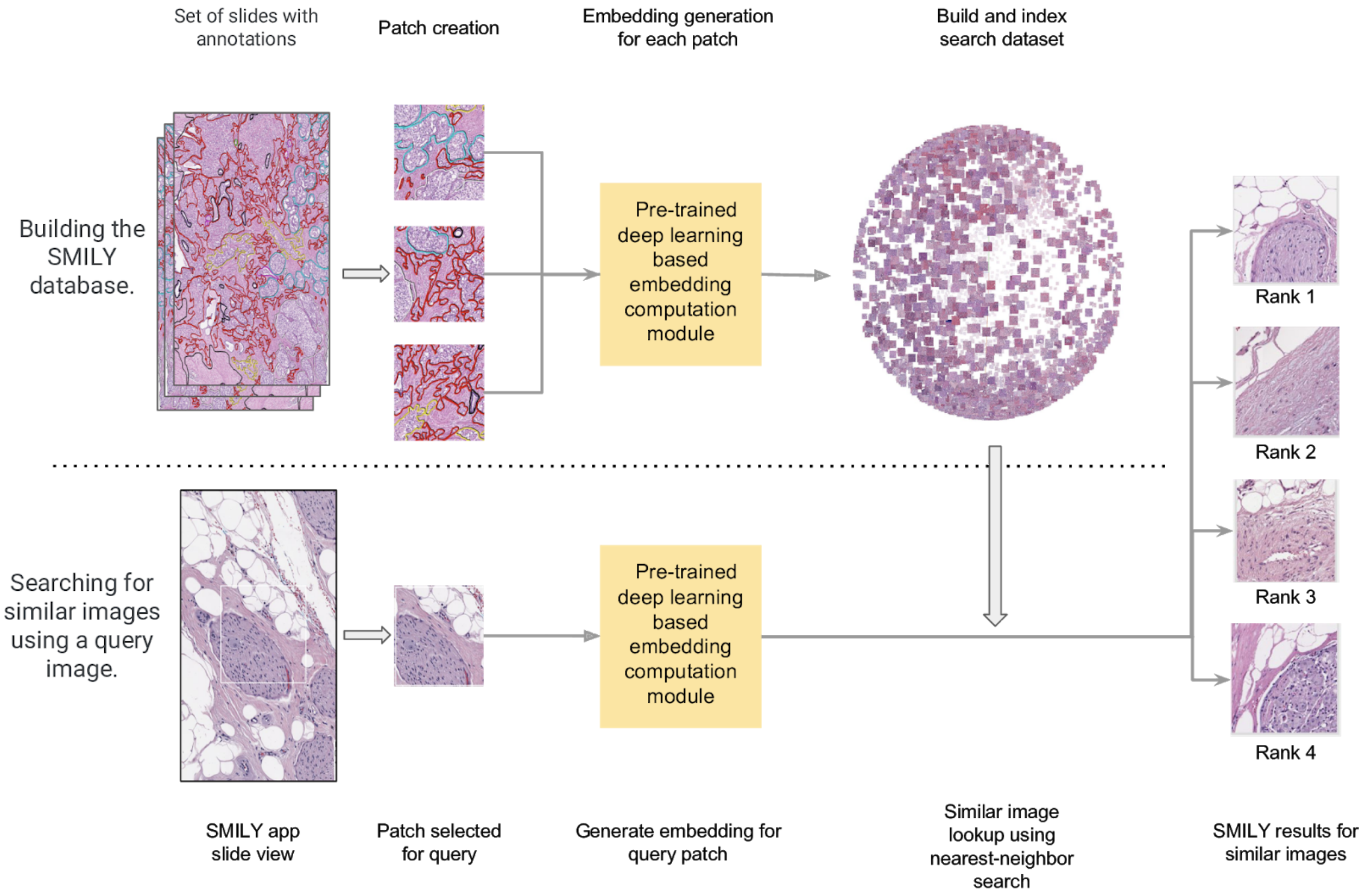

**Fig. 1 | Overview of Similar Medical Images Like Yours (SMILY).** First, a database of image patches and a numerical characterization of each patch's image contents (termed the embedding) is created. SMILY uses a convolutional neural network to compute this embedding (schematic used for illustration purposes only, see Methods for architecture descriptions). Next, when a query image is selected, SMILY computes the embedding of that query image and compares the embedding with those in the database in a computationally efficient manner. Finally, SMILY returns the k most similar patches, where k is customizable.

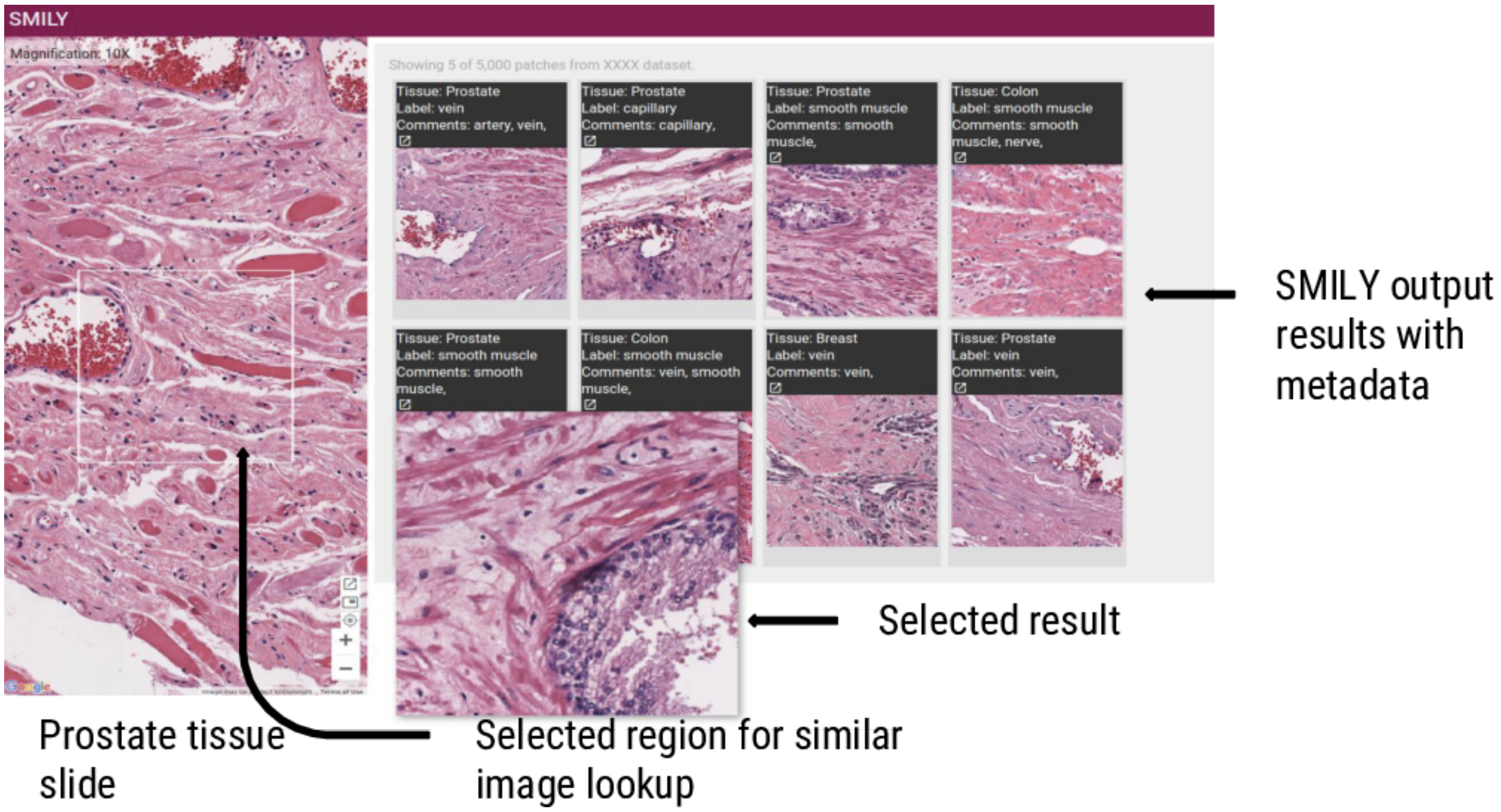

**Fig. 2 | Sample view of the SMILY user interface.** Sample query from a prostate specimen and search results. One of the search results has been magnified for better visualization. Additional examples of queries and search results are presented in Supp Fig. 3, including an additional interface for scoring the quality of each search result for the prospective studies with pathologists.

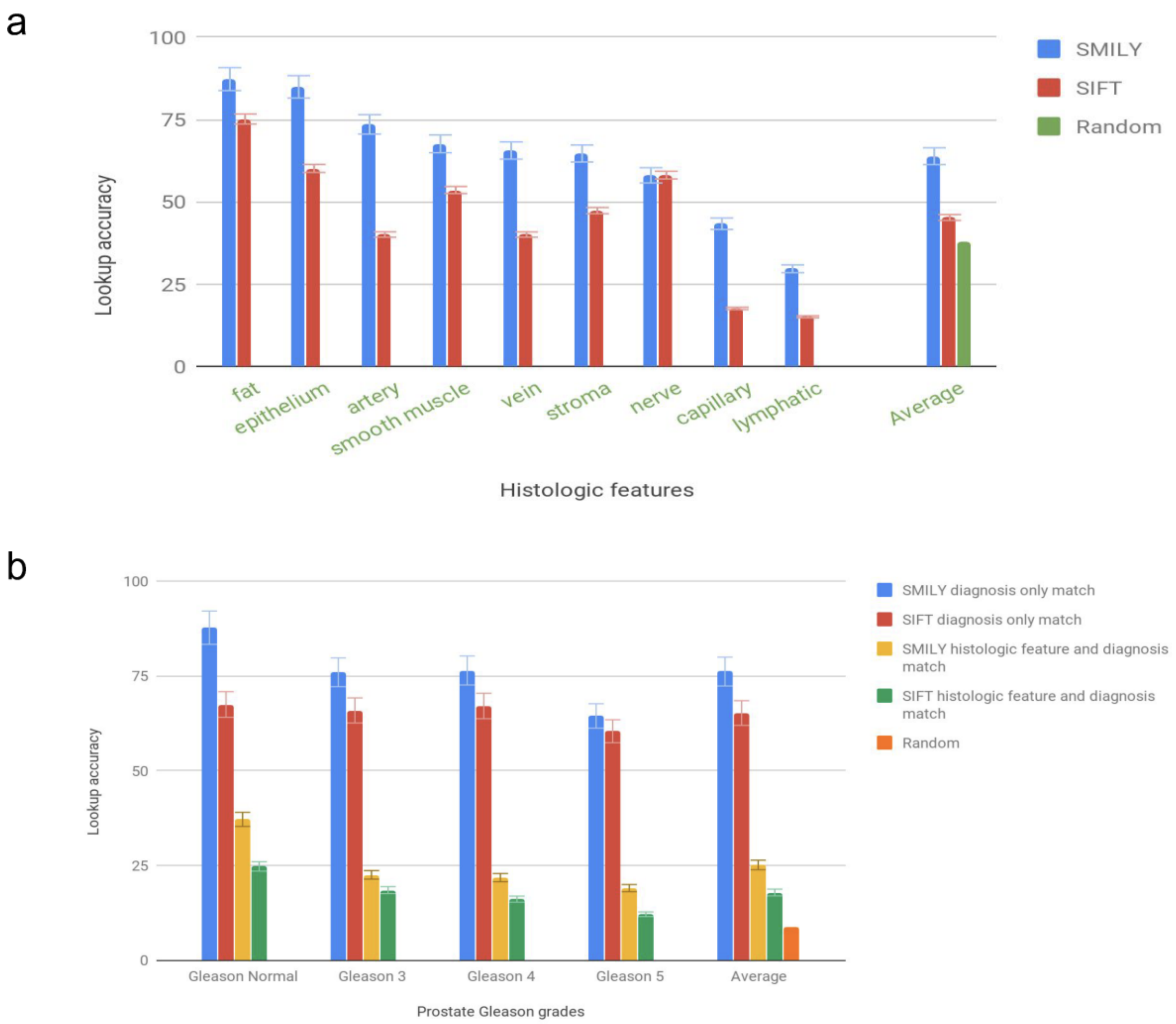

**Fig. 3 | SMILY search accuracy from large-scale quantitative evaluation using pathologist-provided annotations. (a)** Results for histologic feature match in prostate specimens, in comparison with a traditional image feature extractor (scale-invariant feature transform, SIFT) and random search. **(b)** Results for prostate cancer Gleason grade and histologic feature match, in comparison with the same baselines.

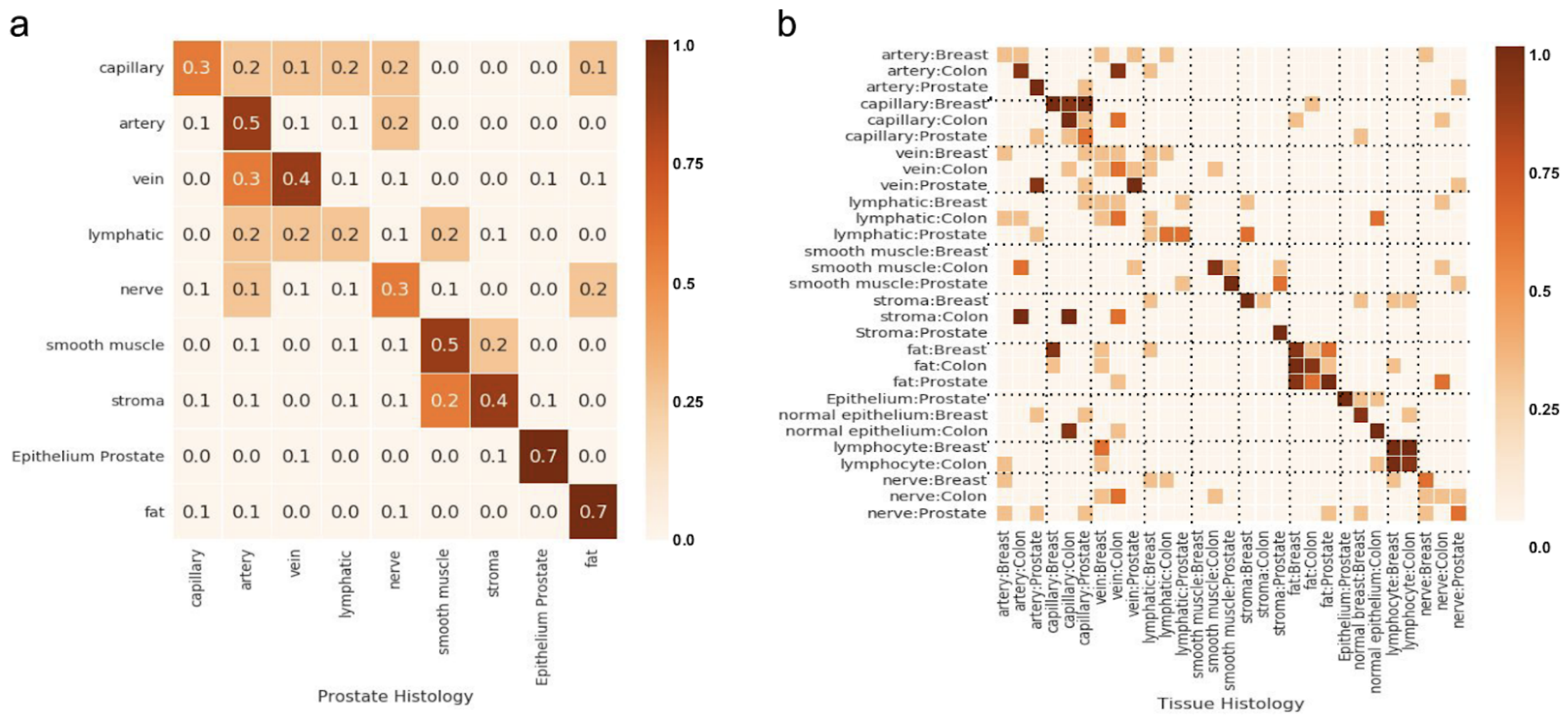

**Fig. 4 | Confusion matrix from SMILY search.** An element in row *i*, column *j* indicates the fraction of search results for query *i* that result in a “hit” based on the top-5 score for the category *j*. **(a)** Confusion matrix for the results from Fig. 3a: histologic feature match in prostate specimens. **(b)** Confusion matrix for histologic feature match across prostate, breast, and colon specimens. To improve visual contrast and highlight trends better, only discrete colors and rounded-off values are used.

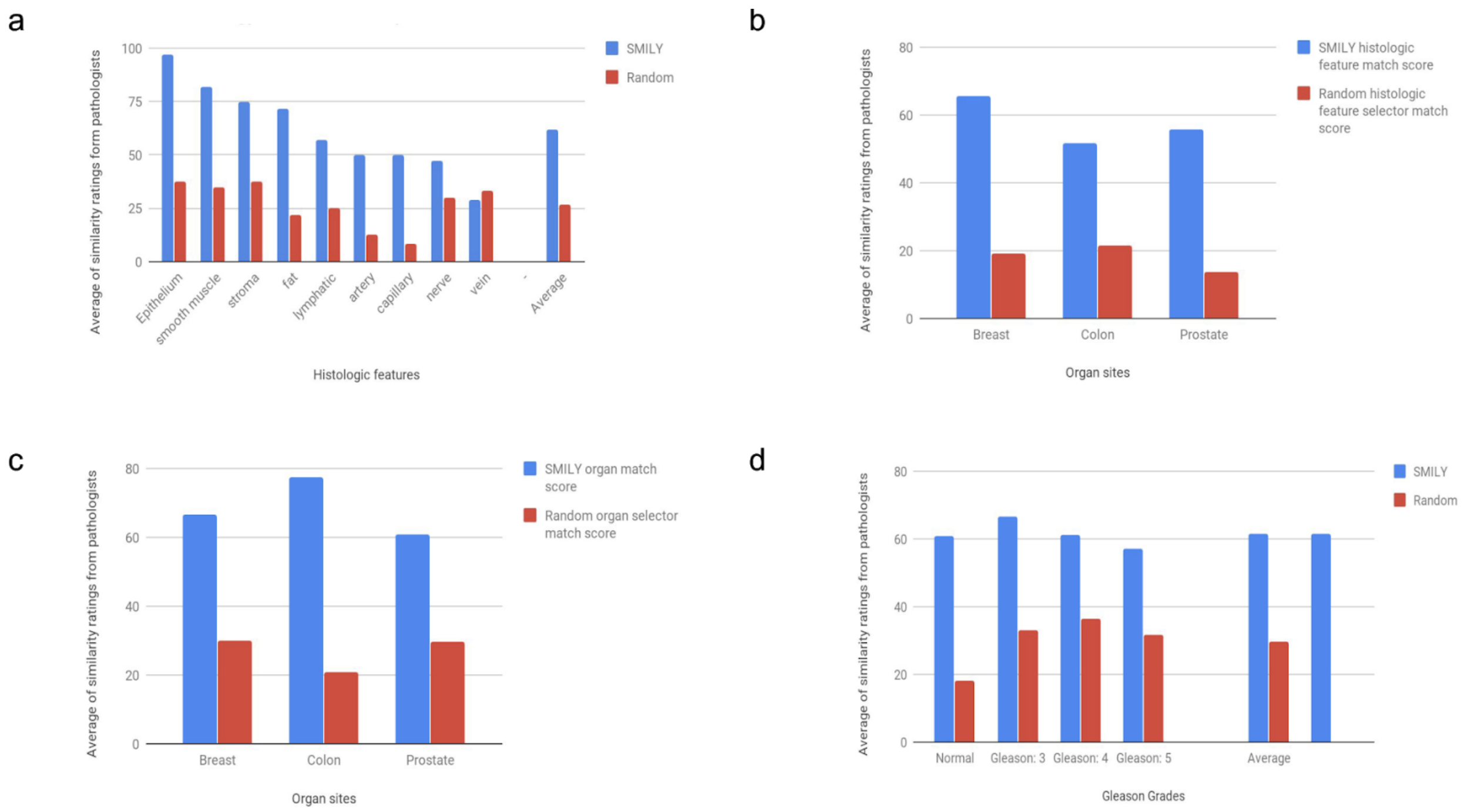

**Fig. 5 | Evaluation of SMILY from studies with pathologists.** The pathologists evaluated search results, blinded to whether the results were retrieved by SMILY versus a negative control, random selection. **(a)** Histologic feature match in prostate specimens. **(b)** Histologic feature match in prostate, breast, and colon specimens. **(c)** Organ site match in prostate, breast, and colon specimens. **(d)** Overall match score (Methods, Table 3) in prostate specimens for similarity in histology and prostate cancer Gleason grade.

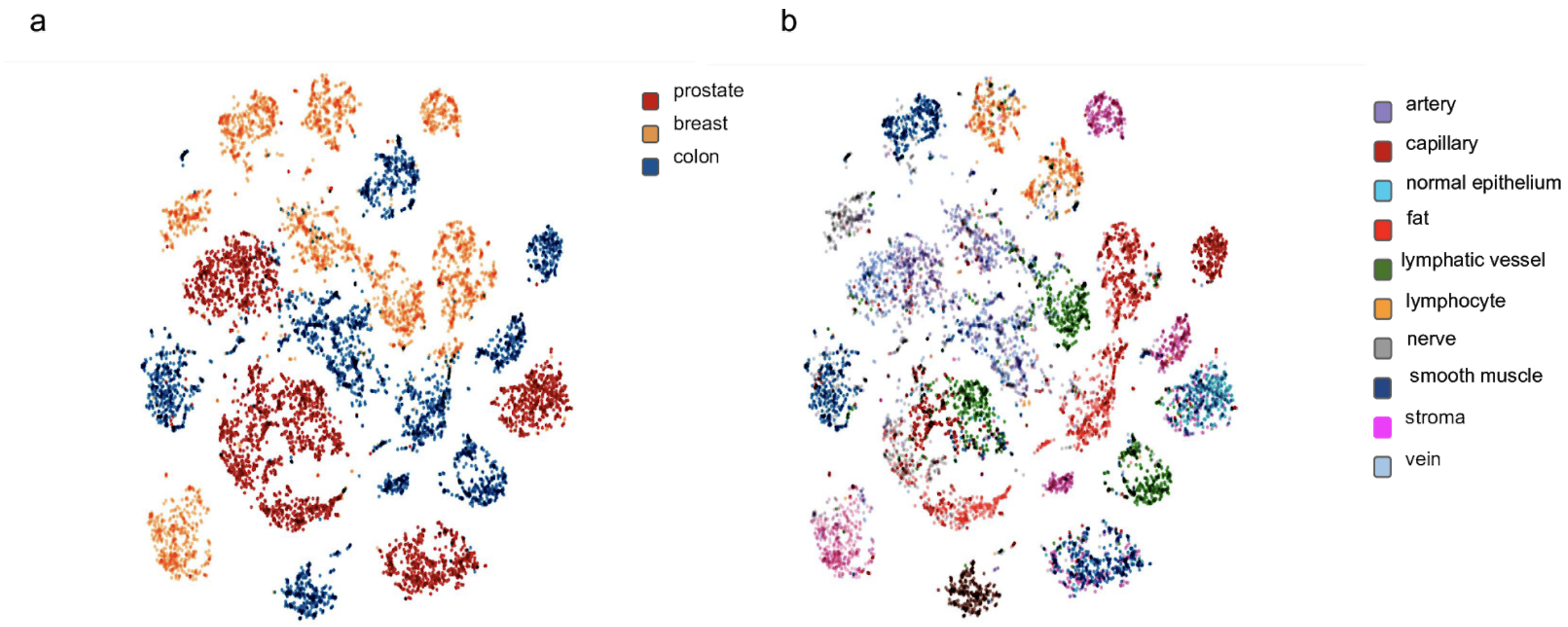

**Fig. 6 | Visualizations of the embeddings of image patches in the SMILY database.** Each dot represents an image patch. **(a)** Colored by organ site, indicating that patches from the same organ were distributed among different clusters. **(b)** Colored by histologic feature, indicating a more distinct separation between histologic features.

# Tables

**Table 1 | Summary of data used in large-scale quantitative study.** To avoid biases in the evaluation, we randomly subsampled the original annotated regions, resulting in 5,000 patches per histologic feature per organ.

| Dataset | Organ site(s) | Categories assessed | Database | | Query set | |
|---|---|---|---|---|---|---|
| | | | Number of slides | Number of patches | Number of slides | Number of patches |
| Organ-specific | Prostate | 9 histologic features | 20 | 45,000 (5,000 per feature) | 5 | 9,000 (1,000 per feature) |
| Multi-organ | Prostate, breast, colon | 10 histologic features | 60 | 87,000 (3,000 per feature/organ*) | 15 | 14,500 (500 per feature/organ*) |
| Gleason grading | Prostate | Non-tumor and Gleason Patterns 3,4,5 (NT, GP3, GP4, GP5) | 20 | 40,000 (10,000 in each category) | 5 | 8,000 (2,000 in each category) |

*In our study, no lymphocytes were found upon non-exhaustive review of the prostate specimens, so the number of patches exclude this.

**Table 2 | Summary of data used in the studies with pathologists.** The database used for this study is identical to Table 1, while the query set was subsampled to retain a tractable number for manual evaluations.

| Dataset | Organ site | Categories assessed | No. of patches for query | | Scoring system (see Methods) |
|---|---|---|---|---|---|
| | | | SMILY | Random (negative control) | |
| Organ-specific | Prostate | 9 histologic features | 270 by 2 pathologists | 90 by 2 pathologists | 0 or 100 for histologic feature match |
| Multi-organ | Prostate, breast, colon | 10 histologic features | 410 by 2 pathologists | 145 by 2 pathologists | 2 scores: 0 or 100 for histologic feature match, and 0 or 100 or “unclear” for organ match |
| Gleason grading | Prostate | Non-tumor and Gleason Patterns 3,4,5 (NT, GP3, GP4, GP5) | 250 by 2 pathologists | 120 by 2 pathologists | 0 to 100 for tumor grade and histologic feature match |

**Table 3 | Overall match quality score for multi-aspect similarity evaluation.**

| Score | Criteria |
|---|---|
| 0 | If the presence/absence of tumor in both patches don't match and they look visually different |
| 25 | If the presence/absence of tumor in both patches don't match and but histological features match |
| 50 | If the presence of tumor in both patches match but not the tumor grade |
| 75 | If the diagnostic grades match or both patches are normal (ex: Gleason grade for prostate) |
| 100 | If the diagnostic grades match or both patches are normal (ex: Gleason Grades for Prostate) in addition to at least one histological feature match |

## Supplementary Material

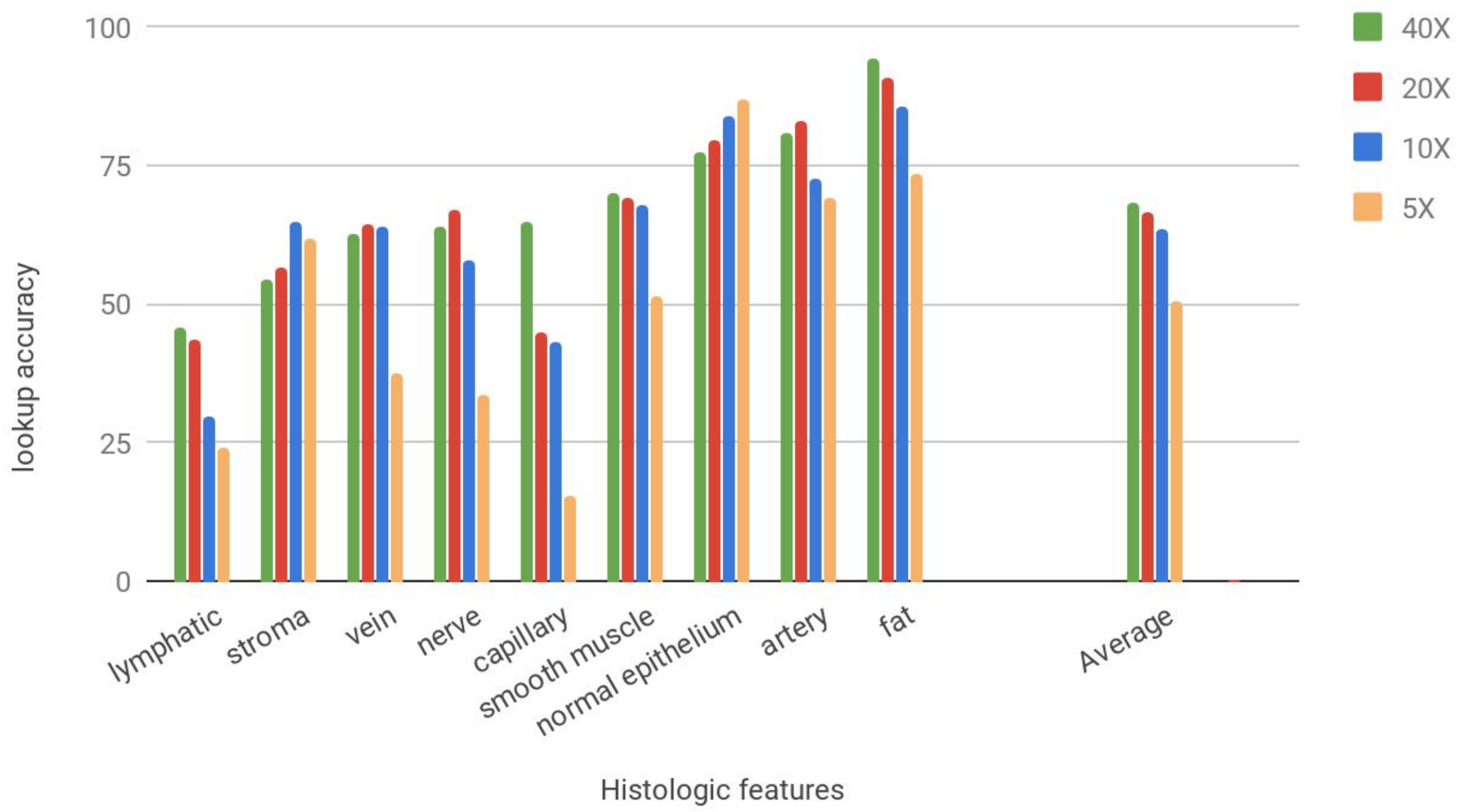

**Supp. Fig. 1 | Top-5 score for histology match using SMILY at different magnifications.** The 5X magnification results are lower because there are substantially fewer image patches per slides at that magnification, of which even fewer pass the 1,000-pixel filter used to enhance search result diversity.

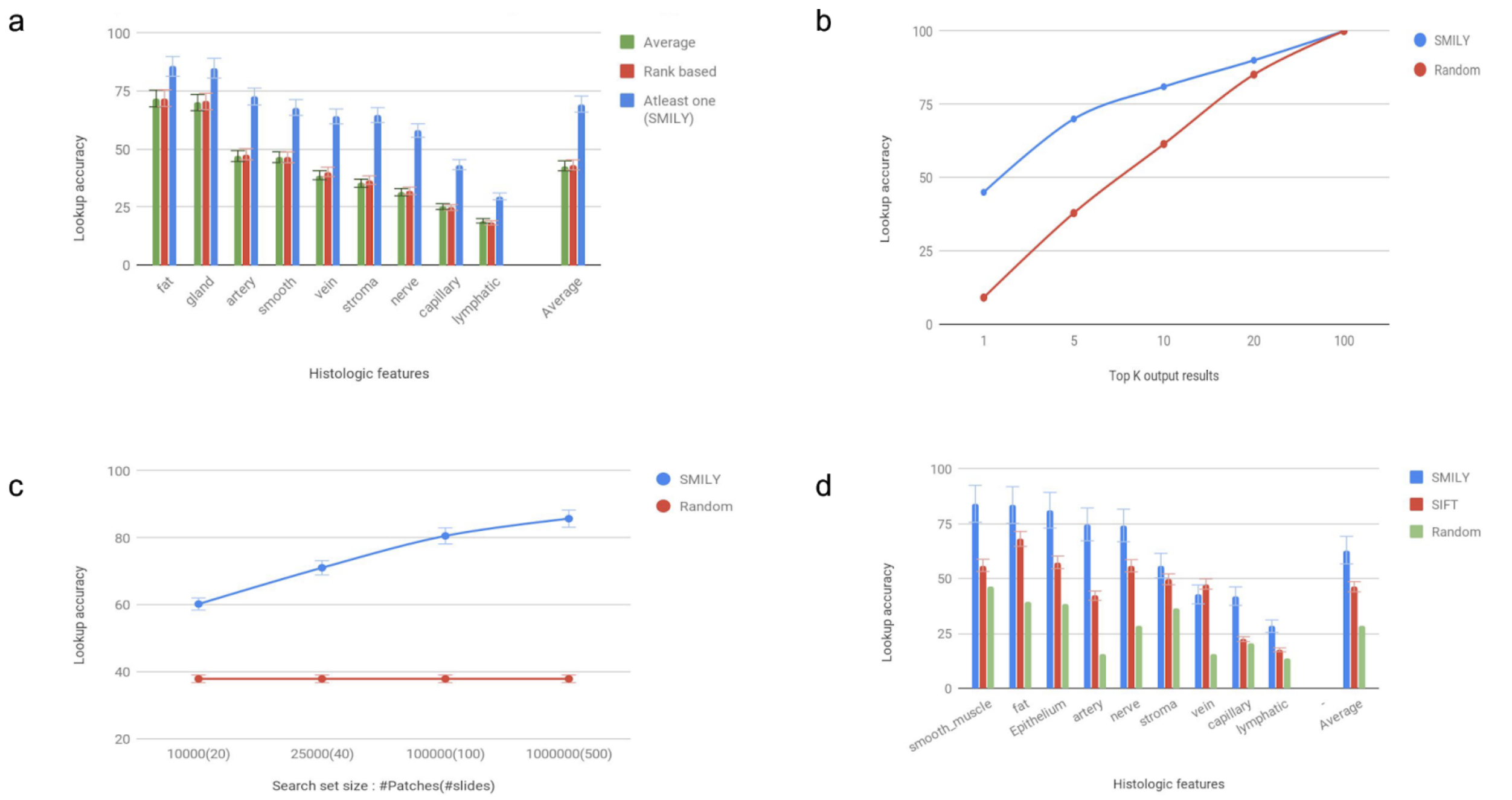

**Supp. Fig. 2 | Additional performance metrics for SMILY. (a)** Comparison different metrics: averaging the histology match (1/0), weighted based on rank in the search results, and at-least-one (top-5 score). **(b)** Comparison of different values of "k" in the top-k score (results for histologic feature match in prostate specimens). **(c)** Effect of search database size (results for gleason cancer grade match in prostate specimens). **(d)** Top-5 score when searching the original annotations in prostate specimens (without resample to ensure uniform distribution of histologic features).

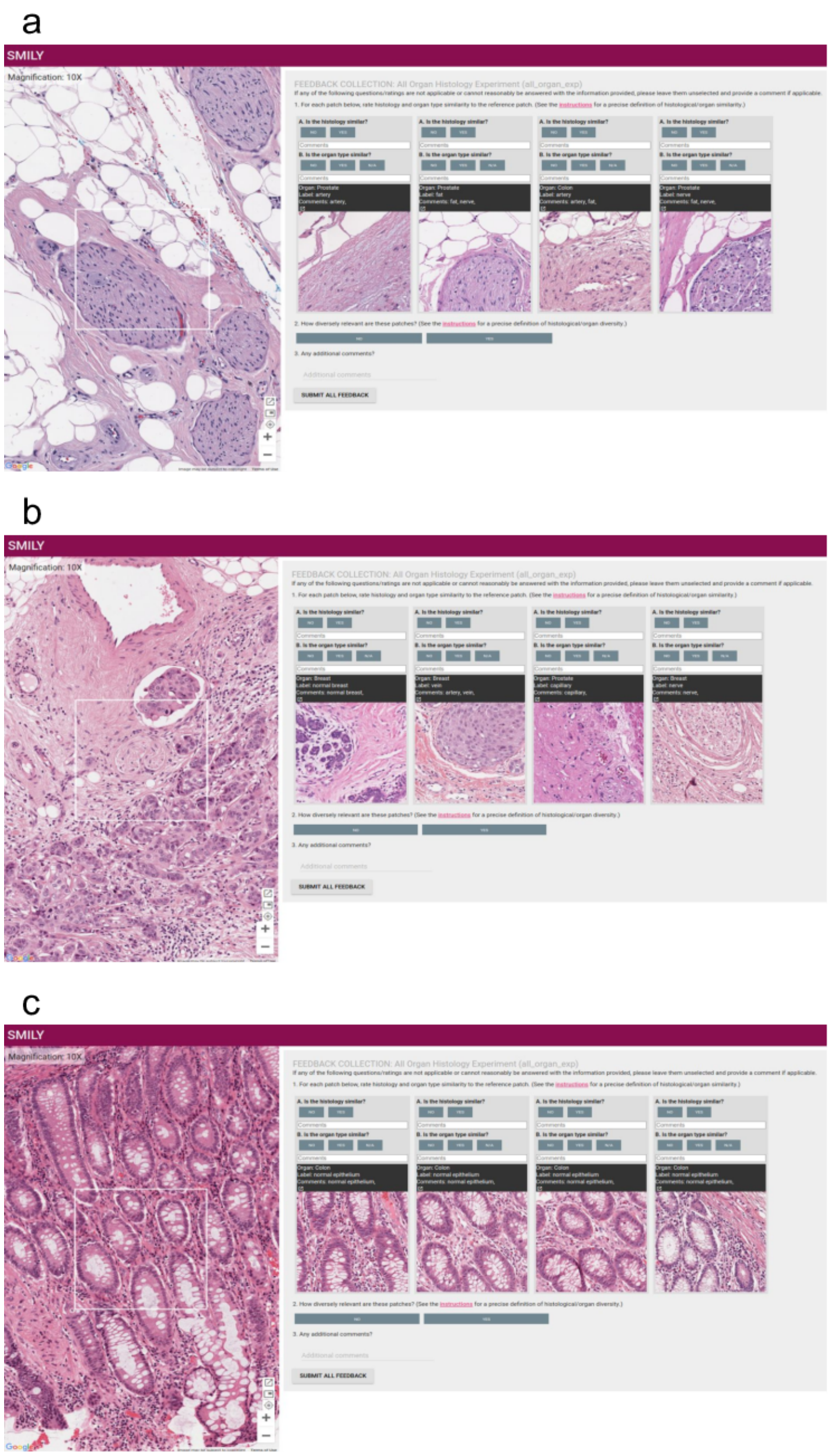

**Supp Fig. 3 | Samples of SMILY similar image retrieval (a)** Query consists of a nerve from a prostate specimen, alongside stromal connective tissue, adipocytes, and blood vessels. The search results contained (1) connective tissue adjacent to a blood vessel, (2) a nerve adjacent to fat, (3) a blood vessel adjacent to fat, and (3) a nerve adjacent to fat and connective tissue. **(b)** Query consists of nerve, stroma, tumor, and a lymphatic vessel containing metastatic tumor from a breast specimen. The search results contained (1) a large tumor focus and stroma, (2) a nerve adjacent to stroma and muscle connective tissue, (3) a lymphatic vessel with stroma, and (4) a nerve adjacent to stroma and a blood vessel. **(c)** Query consists of colonic glands cut in cross and longitudinal sections. The search results all contained colonic glands with similar architecture.

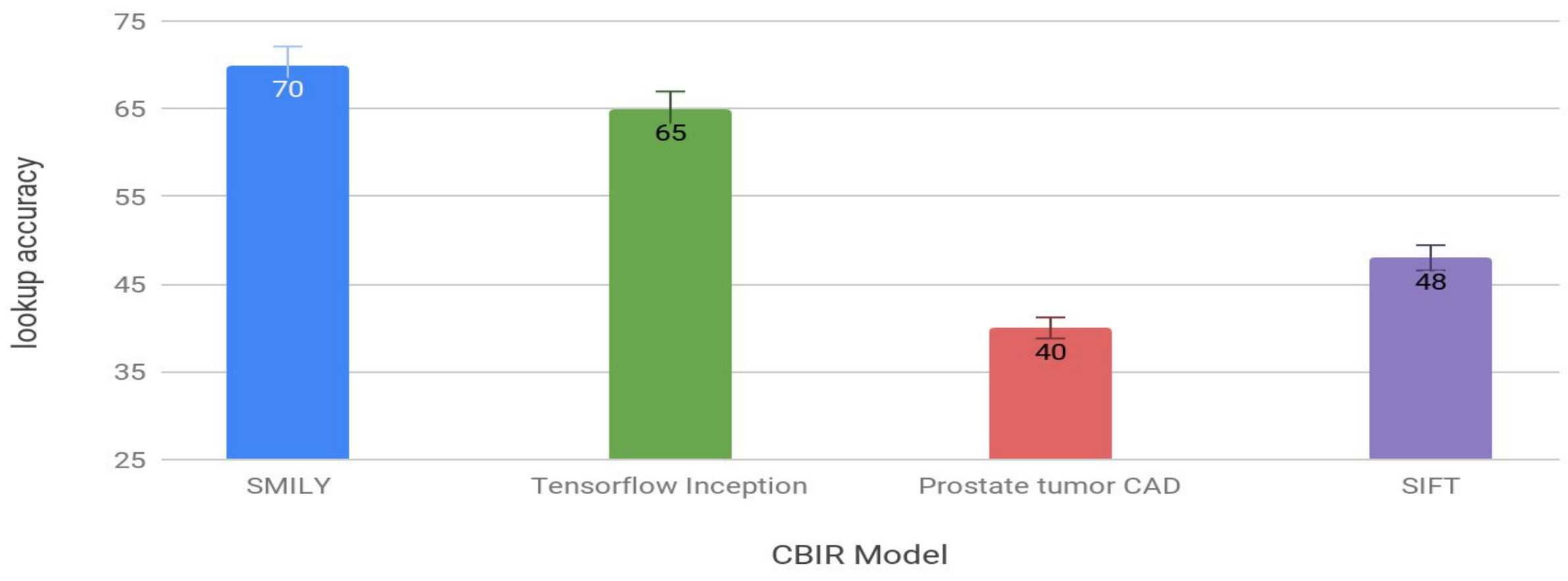

**Supp. Fig. 4 | Comparison of using different pre-trained embedding generators:** SMILY's deep ranking network, Inception (V3)[1,2], a network trained for Gleason grading[3], and scale-invariant feature transform (SIFT)[4].

## Supplemental References